\documentclass[conference]{IEEEtran}
\ifCLASSINFOpdf
\else
\fi
\usepackage{amsmath}
\usepackage{amsfonts}
\usepackage{amsthm}

\usepackage[hidelinks]{hyperref}
\usepackage{booktabs}
\usepackage{multirow}
\usepackage{makecell}
\usepackage{graphicx}
\usepackage{subcaption}
\usepackage{array}

\begin{document}
%
\title{Does Graph Compression Preserve Signal Propagation?}

\author{

\IEEEauthorblockN{Kawshik Banerjee and Khaled Mohammed Saifuddin}
\IEEEauthorblockA{School of Computing, Southern Illinois University, Carbondale, IL, USA}
\IEEEauthorblockA{kawshik.banerjee@siu.edu, k.saifuddin@siu.edu}

}


%


\maketitle

\begin{abstract}
Graph compression reduces the computational cost of graph learning, but its effect on signal propagation remains largely underexplored. Existing work evaluates compression through downstream task performance or structural preservation, neither of which directly captures how propagation dynamics change after compression. We study two fundamental compression paradigms, coarsening and sparsification, and ask whether they preserve the propagation behavior of the original graph. Across five datasets, varying compression rates, and propagation depths, we measure signal behavior through three complementary metrics. Our results reveal a consistent tension between the two compression families. Sparsification retains higher signal diversity and mitigates oversmoothing, but its propagation trajectory progressively diverges from that of the original graph. Coarsening more faithfully preserves propagation behavior, but at the cost of stronger smoothing and rank collapse. These findings demonstrate that two propagation-centric objectives, preserving signal diversity and preserving propagation fidelity, are distinct and empirically at odds under graph compression, highlighting the need for evaluation protocols that jointly consider both dimensions. The code and results are available at: \href{https://github.com/KawshikBanerjee/Compression-Propagation-Duality}{https://github.com/KawshikBanerjee/Compression-Propagation-Duality}

\end{abstract}



%
\IEEEpeerreviewmaketitle

\section{Introduction}
Graph compression has emerged as a practical solution to the computational and memory demands of large-scale graph learning, reducing graph size while aiming to retain properties necessary for effective learning \cite{bravo2019unifying, duan2022comprehensive, liu2023dspar, xu2025learning}. Graph compression research has converged on two dominant paradigms: coarsening, which merges nodes into supernodes to reduce graph order \cite{loukas2019graph, loukas2018spectrally}, and sparsification, which prunes edges to reduce graph size \cite{zhang2025graph, lindner2015structure}.

Despite substantial progress in compression methodology, evaluation has centered on a narrow set of downstream proxies. The predominant approach assesses compression quality through downstream task performance, such as classification accuracy or link prediction area under the curve (AUC) \cite{yang2023does, gc4nc}. A complementary line of work evaluates structural preservation, examining whether spectral properties \cite{kataria2024ugc} or cut guarantees \cite{loukas2019graph} are retained after compression. While both perspectives are valuable, neither directly addresses whether compression preserves how node features evolve under repeated application of the graph operator \cite{gcn}. We refer to this as signal propagation throughout the paper.

This distinction has practical consequences. Propagation is the mechanism through which graph structure influences node representations. If compression alters propagation dynamics, then the signal a node receives after compression differs fundamentally from what it would receive on the original graph, regardless of whether downstream task metrics appear unaffected \cite{joly2024graph}. Understanding this gap has direct implications for when and how compression can be safely applied.

In this work, we take a propagation-centric view of graph compression. We evaluate representative coarsening and sparsification methods across multiple graph datasets at varying compression rates and propagation depths, measuring signal behavior through complementary metrics that capture both intrinsic smoothness and fidelity to the original propagation trajectory. Our analysis reveals a consistent and sharp tension between the two paradigms. Coarsening methods closely follow the original propagation trajectory, but at the cost of stronger signal smoothing, i.e., oversmoothing \cite{cai2020note, oonograph, rusch2023survey}. Sparsification methods better preserve signal diversity and resist oversmoothing, yet their propagation trajectories diverge progressively from the original graph as depth increases. These two objectives, preserving signal diversity and preserving propagation fidelity, are empirically at odds under both paradigms.

Our main contributions are as follows:
\begin{itemize}
    \item We identify and empirically characterize signal diversity preservation and propagation fidelity as two distinct and empirically competing evaluation objectives under graph compression.
    \item We provide a systematic empirical analysis of six compression methods across two paradigms, five datasets, multiple compression rates, and propagation depths using three propagation-centric metrics.
    \item We show that the coarsening-sparsification tension generalizes across methods within each paradigm, with important within-category variation, and demonstrate that methods favorable under oversmoothing-oriented metrics may simultaneously exhibit large deviations from the original propagation trajectory.
\end{itemize}

\section{Related Work} 

Graph compression reduces a large graph to a smaller one while preserving properties relevant to downstream tasks \cite{xu2025learning, hashemi2024comprehensive}. Rather than evaluating compression by its effect on task performance, we study how the propagation operator itself behaves under compression — how closely the signal propagated on a compressed graph reproduces that of the original. We focus on two paradigms that admit a training-free propagation operator: coarsening, which reduces the node set, and sparsification, which reduces the edge set.

\paragraph{Graph Coarsening}
Graph coarsening reduces a graph by grouping nodes into supernodes and aggregating their connections \cite{loukas2018spectrally}. Heavy Edge Matching (HE) \cite{karypis1998fast} iteratively matches adjacent node pairs by the heaviest edge and contracts them into single supernodes, prioritizing strong local connectivity. Variation Neighborhoods (VN) \cite{loukas2019graph} takes a broader view by grouping each node with all its neighbors as candidate contraction sets, and greedily contracts the group with the smallest local variation — a localized measure of how non-smooth the preserved signals are across the group. Both methods rely on pairwise or local structural similarity as the merging criterion. NOPE \cite{nope} departs from this by prioritizing collective neighborhood interference: rather than optimizing each merge independently, it penalizes neighborhood-level deviation, avoiding merging decisions that preserve pairwise similarity at the cost of disrupting the surrounding semantic structure.

\paragraph{Graph Sparsification}
Graph sparsification approximates a dense graph with a sparse one by removing a subset of edges \cite{fung2011general}, motivated by the theoretical guarantee that every graph has a spectral sparsifier that is nearly-linear in size \cite{spielman2011sparsifier}. In practice, sparsification methods assign a score to each edge based on some structural objective and apply global filtering to retain a target fraction of edges \cite{lindner2015structure}. Random Edge (RE) \cite{lindner2015structure} is the simplest approach, uniformly sampling edges without regard to graph structure. Local Degree (LE) \cite{lindner2015structure} improves on this by retaining edges incident to the highest-degree neighbors of each node, thereby preserving local connectivity and hub structure, following community detection approaches \cite{satuluri2011local}. TEDDY \cite{teddy} takes a more principled approach, assigning degree-based scores to edges and pruning high-degree edges in a single pass while retaining low-degree ones that serve as structurally critical links between otherwise distant parts of the graph.

\paragraph{Oversmoothing and Propagation Under Compression}
Oversmoothing is a well-documented phenomenon in graph learning, where node representations become increasingly similar with depth, eventually collapsing into near-identical vectors \cite{li2018deeper, oonograph}. Dirichlet energy is commonly used to measure this effect and has been shown to decay exponentially with depth \cite{oonograph, cai2020note, wu2023demystifying}. Recent work questions the sufficiency of energy-based metrics, showing that performance degradation can precede meaningful energy decay, and proposes rank-based alternatives \cite{zhang2026we}. Since compression alters the propagation operator itself, oversmoothing and compression are not independent: \cite{cai2020note} shows that dropping edges and coarsening nodes shift a graph's eigenvalues and Dirichlet energy in similar ways. Closer to our setting, \cite{joly2024graph} shows that spectral preservation under coarsening does not guarantee that message passing on the coarsened graph matches the original and proposes a new propagation matrix with explicit message-passing guarantees.

\paragraph{Positioning of This Work}
Compression is evaluated in the literature mostly through downstream task performance, while the oversmoothing literature characterizes how propagation degrades feature representations independently of compression. We connect these threads, asking whether signal diversity preservation and propagation fidelity are distinct objectives under compression, and whether the two compression categories sit on complementary sides of this trade-off.

\section{Problem Setup and Methodology}

\subsection{Overview}
We investigate how different graph compression paradigms alter signal propagation dynamics. Specifically, we contrast two fundamentally different compression categories, coarsening and sparsification, and ask whether they preserve the propagation behavior of the original graph. To this end, we employ three complementary metrics that jointly capture intrinsic signal smoothness and fidelity to the original propagation trajectory.

\subsection{Baseline Propagation}
To study how compression alters signal propagation, we first establish a reference propagation on the original uncompressed graph. This baseline captures how node features spread through the graph structure under repeated neighborhood aggregation and serves as the ground truth against which compressed graph propagation is compared.

We consider a graph $\mathcal{G} = (\mathcal{V}, \mathcal{E})$ where $\mathcal{V}$ denotes a set of $N$ nodes and $\mathcal{E}$ denotes a set of $E$ edges. We define the baseline propagation operator following the symmetric normalized adjacency matrix with self-loops introduced in \cite{gcn}: $ \tilde{A} = \tilde{D}^{-\frac{1}{2}} (A+I) \tilde{D}^{-\frac{1}{2}} $ where $A$ is the adjacency matrix, $I$ is the identity matrix, and $\tilde{D}$ is the degree matrix of $A+I$. Given the initial node feature matrix $X \in \mathbb{R}^{N \times d}$ of $d$-dimensional node features, the propagated signal at depth $k$ is:
\begin{equation}
\label{eq:baseline}
    Y^{(k)} = \tilde{A}^k X.
\end{equation}
Note that we do not train any model; $\tilde{A}$ is used solely to measure how signals evolve through the graph structure at increasing depths.

\subsection{Graph Compression}
Graph compression seeks to reduce the size of a graph $\mathcal{G} = (\mathcal{V}, \mathcal{E})$ while retaining its structural properties. We contrast two fundamentally different compression paradigms: coarsening and sparsification. In both cases, we treat compression as an operation that alters the graph topology, and we evaluate its effect on signal propagation rather than on trained model performance.

\subsubsection{Graph Coarsening}
Graph coarsening produces a smaller graph $\mathcal{G}_c = (\mathcal{V}_c, \mathcal{E}_c)$ by partitioning the original node set $\mathcal{V}$ into $N'$ disjoint groups and merging each group into a single supernode, where $\lvert \mathcal{V}_c \rvert = N' < N$. Formally, a cluster assignment matrix $C \in \{0, 1\}^{N' \times N}$ encodes this partition, where $C_{\mu i} = 1$ if and only if node $i$ belongs to supernode $\mu$. The initial features in $X$ are averaged within each group to form the coarsened feature matrix $X_c \in \mathbb{R}^{N' \times d}$, and the coarse propagation operator $\tilde{A}_c$ is constructed on the resulting supernode graph. The coarse signal at depth $k$ is then: $ Y_c^{(k)} = \tilde{A}_c^k X_c $.

Since the coarsened graph operates in a reduced node space, direct comparison with the original baseline is not straightforward. We therefore map the coarse signal back to the original node space by applying the transpose of $C$:
\begin{equation}
\label{eq:coarse_signal_lifted}
    Y_m^{(k)} = C^\top Y_c^{(k)}.
\end{equation}

\subsubsection{Graph Sparsification}
Graph sparsification produces a sparser graph $\mathcal{G}_s = (\mathcal{V}, \mathcal{E}_s)$ by removing a subset of edges from $\mathcal{E}$, where $\mathcal{E}_s \subseteq \mathcal{E}$, while preserving the original node set $\mathcal{V}$. Since the node set remains unchanged, the sparsified signal can be compared directly against the original baseline without any lifting step. The sparsified graph yields a modified propagation operator $\tilde{A}_s$, and the signal trajectory is measured as:
\begin{equation}
\label{eq:sparse_signal}
    Y_s^{(k)} = \tilde{A}_s^k X.
\end{equation}
\subsection{Evaluation Metrics}
We assess the effect of compression on signal propagation through three complementary metrics, each capturing a distinct aspect of propagation behavior. We report these metrics across multiple propagation depths to capture both early and asymptotic propagation regimes. Shallow depths reflect the immediate effect of compression on local neighborhood aggregation, while deeper depths reveal whether compressed graphs exhibit qualitatively different long-range behavior, such as accelerated rank collapse or sustained deviation from the original trajectory.

\subsubsection{Dirichlet Energy Ratio}
Dirichlet energy (DE) measures the total variation of a signal across graph edges and is one of the most widely used metrics to quantify oversmoothing \cite{rusch2023survey}. Intuitively, as propagation depth increases, repeated neighborhood aggregation causes node representations to become increasingly similar, driving the energy toward zero. Following \cite{cai2020note, zhou2021dirichlet}, we measure the Dirichlet energy of the signal $Y^{(k)} = \tilde{A}^k X$ at depth $k$ as:
\begin{equation}
    E^{(k)} = \mathrm{tr}\!\big((Y^{(k)})^\top L\, Y^{(k)}\big) = \tfrac{1}{2}\sum_{i,j} A_{ij}\,\lVert y_i^{(k)} - y_j^{(k)}\rVert_2^2
\end{equation}
where $L$ is the unnormalized graph Laplacian, and a lower $E^{(k)}$ indicates stronger oversmoothing \cite{zhou2021dirichlet, zhou2025tined}. Since raw energy values depend on input feature scale and graph size, comparing them directly across compression categories is not meaningful. Therefore, following \cite{zhou2025tined}, we normalize by the initial energy at $k=0$ and measure the DE ratio $R^{(k)}$:
\begin{equation}
    R^{(k)} = \frac{E^{(k)}}{E^{(0)}}.
\end{equation}
A lower $R^{(k)}$ indicates stronger smoothing relative to the original signal energy, with a ratio approaching zero implying near-complete oversmoothing.

\subsubsection{Deviation}
While the energy ratio measures the intrinsic smoothness of the compressed signal, it does not capture whether the compressed graph preserves the propagation behavior of the original graph. Two graphs could exhibit identical energy decay profiles yet produce entirely different signal trajectories. To capture this, motivated by \cite{joly2024graph}, we measure the relative Frobenius error ($\Delta_k$) between the compressed and original signals at each depth $k$:
\begin{equation}
    \Delta_k = \frac{\left\| Y^{(k)} - Y_{\text{cmp}}^{(k)} \right\|_F}{\left\| Y^{(k)} \right\|_F}, \qquad Y_{\text{cmp}}^{(k)} = \begin{cases} Y_s^{(k)} & \text{sparsification} \\ Y_m^{(k)} & \text{coarsening} \end{cases}
\end{equation}
following equations \ref{eq:baseline}, \ref{eq:coarse_signal_lifted}, and \ref{eq:sparse_signal}. A higher $\Delta_k$ indicates that the compressed graph produces signals that diverge more strongly from the original propagation trajectory.

\subsubsection{Rank}
Rank-based metrics provide a complementary perspective on oversmoothing that is distinct from energy decay. While Dirichlet energy captures the total variation of the signal across edges, rank directly measures the dimensionality of the signal subspace and is a more reliable indicator of oversmoothing in certain cases \cite{zhang2026we}. When oversmoothing occurs, node representations collapse toward a low-dimensional subspace, reducing the rank of the signal matrix. We report the numeric rank (NumRank) that approximates the effective number of linearly independent directions in the signal matrix \cite{numRank}:
\begin{equation}
\operatorname{NumRank} = \frac{\| Y_{\text{cmp}}^{(k)} \|_F^2}{\| Y_{\text{cmp}}^{(k)} \|_2^2}, \qquad Y_{\text{cmp}}^{(k)} = \begin{cases} 
    Y^{(k)} & \text{baseline} \\
    Y_s^{(k)} & \text{sparsification} \\ 
    Y_m^{(k)} & \text{coarsening} 
\end{cases}
\end{equation}

where, $\| . \|_F$ and $\| . \|_2$ denote the Frobenius norm and the spectral norm, respectively.

A higher rank indicates greater signal diversity. For coarsening-based methods, both metrics are measured in the original node space after lifting through $C^\top$, making them directly comparable across compression categories.

\subsection{Experimental Setup}

We evaluate six compression methods on five datasets (see Table \ref{tab:dataset_statistics}) across two categories: NOPE \cite{nope}, Heavy Edge Matching \cite{loukas2019graph}, and Variation Neighborhoods \cite{loukas2019graph} for coarsening, and TEDDY \cite{teddy}, Local Degree \cite{lindner2015structure}, and Random Edge \cite{lindner2015structure} for sparsification. Heavy Edge Matching, Variation Neighborhoods, Local Degree, and Random Edge follow the implementations in \cite{gc4nc}. Each method is applied at three compression rates $r \in \{0.3, 0.5, 0.7\}$ in our experiments, where $r$ denotes the fraction of nodes merged for coarsening and the fraction of edges pruned for sparsification. However, for visual clarity, we report only $r \in \{0.3, 0.7\}$ in our figures, while the complete results are available in our codebase. We report propagation depths $k \in \{2, 4, 8, 16, 32\}$.

\begin{table}[htbp]
\centering
\caption{Dataset Statistics}
\label{tab:dataset_statistics}
\setlength{\tabcolsep}{4pt}
\begin{tabular}{lrrrrrrr}
\toprule
\textbf{Dataset} & \textbf{\#Nodes} & \textbf{\#Edges} & \textbf{\#Features} & \multicolumn{1}{>{\raggedleft\arraybackslash}p{1.2cm}}{\hspace{0pt}\textbf{\#Components}} \\
\midrule
Cora \cite{yang2016revisiting}      & 2,708  & 5,278   & 1,433 & 78  \\
Citeseer \cite{yang2016revisiting}  & 3,327  & 4,552   & 3,703 & 438 \\
Pubmed \cite{yang2016revisiting}    & 19,717 & 44,324  & 500   & 1   \\
DBLP \cite{bojchevski2018deep}      & 17,716 & 52,867  & 1,639 & 589 \\
Amazon Computers \cite{shchur2018pitfalls}    & 13,752 & 245,861 & 767   & 314 \\
\bottomrule
\end{tabular}\vspace{-.5cm}
\end{table}

Each method-dataset-rate combination is run over five random seeds $\{0, 1, 2, 3, 4\}$, and we report the mean $\pm$ standard deviation across seeds. The baseline dense graph propagation is computed on a single fixed seed, as it involves no stochastic components. All experiments were conducted on a single NVIDIA RTX PRO 6000 GPU.

\begin{figure*}[htbp]
\centering
\includegraphics[width=\textwidth]{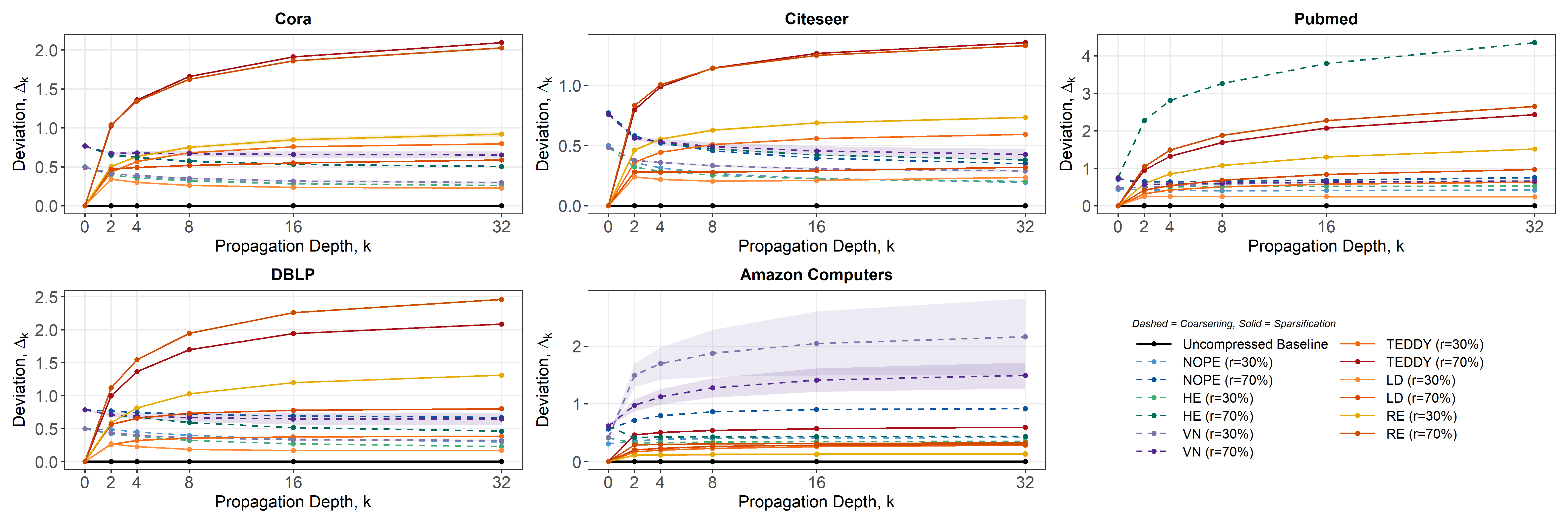}
\caption{Deviation from baseline comparison across compression methods, compression ratios, and datasets.}
\label{fig:deviation}
\end{figure*}

\begin{figure*}[htbp]
\centering
\includegraphics[width=\textwidth]{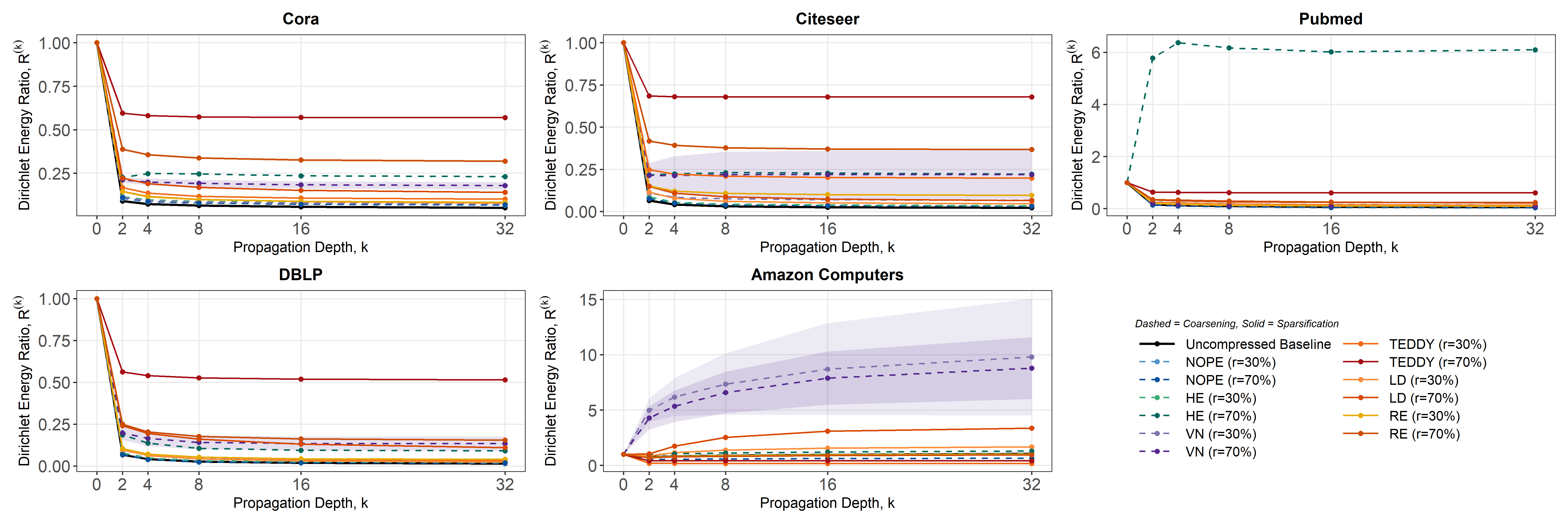}
\caption{Dirichlet energy ratio comparison across compression methods, compression ratios, and datasets.}
\label{fig:decay_ratio}
\end{figure*}

\begin{figure*}[htbp]
\centering
\includegraphics[width=\textwidth]{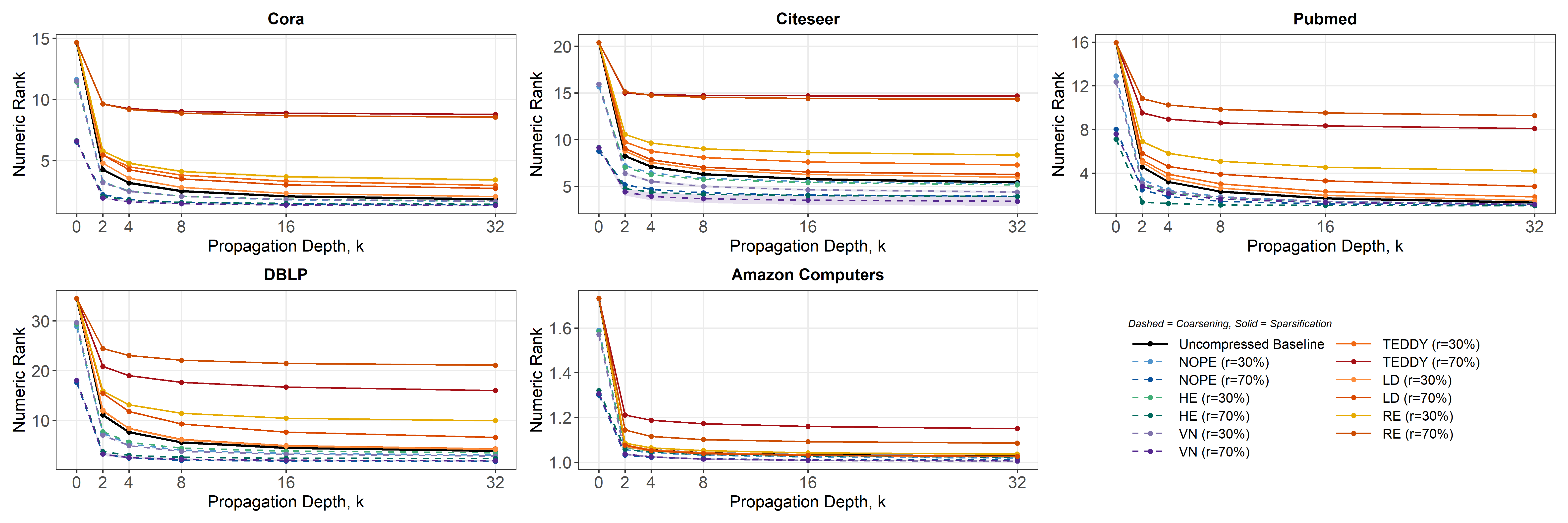}
\caption{Numeric rank comparison across compression methods, compression ratios, and datasets.}
\label{fig:num_rank}
\end{figure*}

\section{Results and Discussion}

Broadly, our findings reveal distinct failure modes for the two compression paradigms. Sparsification-based methods retain higher energy and rank at larger depths than the uncompressed graph, indicating reduced oversmoothing and stronger signal diversity, but at the cost of a propagation trajectory that increasingly diverges from the original as depth increases. Coarsening-based methods show the opposite tendency: they reproduce the original propagation trajectory more faithfully, but generally exhibit stronger oversmoothing, with lower energy and rank at larger depths. This effect is not uniform across settings. Trajectory fidelity and oversmoothing are largely decoupled: a coarsened graph can follow the baseline trajectory closely while its representations are either more or less oversmoothed than the baseline's, depending on dataset structure and propagation depth.

\subsection{Coarsening: Propagation Fidelity at the Cost of Diversity}

Across NOPE, Heavy Edge Matching, and Variation Neighborhoods, deviation $\Delta_k$ consistently decreases with depth on Cora, Citeseer, and DBLP (see Fig \ref{fig:deviation}), indicating that coarsened propagation converges toward the baseline trajectory as depth increases. At the same time, the numeric rank fall below the baseline at all depths and rates, with the gap widening as $r$ increases, consistent with coarsening accelerating oversmoothing through both feature averaging and repeated aggregation on the coarse operator $\tilde{A}_c$.

Pubmed and Amazon Computers break this pattern. On Pubmed, all three methods show deviation that \emph{increases} with depth. Heavy Edge Matching is most extreme (DE ratio exceeding 6 at $r=70\%$, deviation reaching $\sim$4.3 at $k=$32; see Fig. \ref{fig:deviation} and \ref{fig:decay_ratio}), Variation Neighborhoods is milder (between $\sim$0.55 and $\sim$0.75 at $r=70\%$), and NOPE relatively stable (between $\sim$0.65 and $\sim$0.75 at $r=70\%$). We attribute this to cluster size imbalance rather than disconnection: components remain fully connected (singletons $=0$) on Pubmed at all rates, but at $r=70\%$ Heavy Edge Matching's largest cluster contains 697 nodes, compared to 55 for NOPE and 62 for Variation Neighborhoods. A similar pattern is also observed in Amazon Computers as well. Averaging features over large heterogeneous groups produces coarse features that poorly represent individual nodes, causing the coarse operator $\tilde{A}_c$ to amplify rather than smooth the signal. Coarsening's fidelity advantage thus appears conditional on balanced cluster sizes, a condition more easily violated on denser graphs at high compression rates.

\subsection{Sparsification: Signal Diversity at the Cost of Fidelity}
TEDDY, Local Degree, and Random Edge all preserve substantially higher rank than the baseline across all settings (see Fig. \ref{fig:num_rank}), since edge removal does not directly reduce feature dimensionality and slows convergence to a common subspace. DE ratio plateaus above the baseline, with the plateau value increasing monotonically with pruning rate (e.g., TEDDY on Cora rises from $\sim$0.1 at $r=30\%$ to $\sim$0.6 at $r=70\%$ and $k=32$; see Fig. \ref{fig:decay_ratio}).

This comes at a cost of deviation that increases monotonically with depth, the opposite of coarsening. TEDDY's increase is smooth and plateaus ($\sim$2.1 on Cora, $\sim$2.4 on Pubmed, $\sim$0.6 on Amazon Computers at $r=70\%$ and $k=32$; see Fig. \ref{fig:deviation}). Random Edge shows comparable divergence ($\sim$2.0 on Cora, $\sim$2.6 on Pubmed at $r=70\%$ and $k=32$; see Fig. \ref{fig:deviation}), though its graph statistics show 8,980 singletons on Pubmed at $r=70\%$. Since isolated nodes retain their initial features unchanged, its apparent rank advantage can be partly considered an artifact of disconnection.

Local Degree does not fit cleanly into either paradigm: at $r=30\%$, deviation decreases with depth (coarsening-like), while at $r=70\%$ it increases (sparsification-like). On Pubmed, this is most pronounced, with a deviation flattening near 0.25 at $r=30\%$ but rising to $\sim$1.0 at $k=32$ for $r=70\%$. This suggests that its degree-aware criterion preserves enough local structure at low rates to maintain fidelity, but this breaks down as more high-degree-adjacent edges are removed.

\subsection{The Diversity-Fidelity Tension Across Methods}
Together, the six methods support a category-level distinction that is neither absolute nor uniform. Coarsening generally preserves fidelity at the cost of oversmoothing, but this advantage degrades on denser graphs at high compression rates, reversing entirely for Heavy Edge Matching on Pubmed. Sparsification generally preserves diversity at the cost of fidelity, but the magnitude varies with the pruning criterion, from TEDDY's bounded divergence to Random Edge's comparable but partly artifactual divergence inflated by disconnection.

These results suggest that diversity preservation and propagation fidelity reflect two distinct mechanisms: reducing signal dimensionality (coarsening) versus reducing aggregation connectivity (sparsification). A method favorable under oversmoothing-oriented metrics may simultaneously exhibit large, growing deviation from the original trajectory, and vice versa. This has practical implications: applications requiring faithful reproduction of propagation behavior should favor coarsening on sufficiently connected graphs, while applications requiring diverse node representations may favor sparsification despite its growing divergence. However, our analysis is purely propagation-focused and does not connect these findings to downstream task performance. Because $r$ denotes the fraction of nodes merged in one family and the fraction of edges removed in the other, matched ratios represent matched budgets rather than equivalent structural changes. \textit{We leave two directions to future work}: translating propagation fidelity and signal diversity to downstream performance, and defining compression measures that align across graph reduction categories.

\section{Conclusion}
We studied the effect of graph compression on signal propagation, evaluating representative coarsening and sparsification methods across multiple datasets, compression rates, and propagation depths. Our results reveal a consistent tension: coarsening preserves propagation fidelity at the cost of stronger oversmoothing, while sparsification retains higher signal diversity but diverges progressively from the original propagation trajectory. This tension generalizes across methods within each category, with important caveats — aggressive coarsening on denser graphs can destabilize fidelity, and unstructured sparsification inflates rank metrics through disconnection rather than genuine diversity. Together, these findings suggest that signal diversity and propagation fidelity are distinct and competing objectives, and that evaluation frameworks relying solely on oversmoothing-oriented metrics may give an incomplete picture of compression quality.

\section*{Acknowledgement}
Generative AI tools were used for proofreading, sentence enhancement, and grammatical correction in this paper. 

\bibliographystyle{IEEEtran}
\bibliography{references}

\end{document}